\documentclass[onecolumn,11pt,draftcls,journal]{IEEEtran}
\ifCLASSINFOpdf
  \usepackage[pdftex]{graphicx}
\else
\fi
\usepackage{algorithmic}
\usepackage{url}
\usepackage[export]{adjustbox}
\usepackage{amsmath}
\usepackage{rotating}
\usepackage[font=small,labelfont=bf]{caption}
\usepackage{multirow}
\usepackage{array}
\def \dibp {the document image binarization problem\,}
\newcommand{\im}[1]{{\bf{#1}}}
\DeclareMathOperator*{\argmin}{arg\,min}
\DeclareMathOperator*{\argmax}{arg\,max}
\def\fwI{.25}
\def\fwIb{.16}
\def\fwII{.2}

\def\fhI{4.2cm}
\def\fhIb{2.2cm}

\def\citeDIBCO{\cite{DIBCO2009,DIBCO2010,DIBCO2011,DIBCO2012,DIBCO2013,DIBCO2014}\,}
\def\etal {{\textit{et al.}}}

\begin{document}

\hyphenation{op-tical net-works semi-conduc-tor}

%
\title{Learning Document Image Binarization from Data}
\author{ {Yue Wu, Stephen Rawls, Wael AbdAlmageed and Premkumar Natarajan}
\thanks{ Authors are all associated with Information Sciences Institute, 
University of South California, Marina Del Ray, California 90230. Email: \{yue\_wu, srawls, wamageed, pnataraj\}@isi.edu.}}
\maketitle

\setlength{\intextsep}{3pt} 
\setlength{\textfloatsep}{3pt}
\setlength{\abovecaptionskip}{3pt}
\setlength{\belowcaptionskip}{1pt}
\setlength{\dbltextfloatsep}{1pt}

\begin{abstract}
In this paper we present a fully trainable binarization solution for degraded document images. Unlike previous attempts that often used simple features with a series of pre- and post-processing, our solution encodes all heuristics about whether or not a pixel is foreground text into a high-dimensional feature vector and learns a more complicated decision function. In particular, we prepare features of three types: 1) existing features for binarization such as \textit{intensity}  \cite{Otsu1975}, \textit{contrast} \cite{Su2009,Su2013}, and \textit{Laplacian} \cite{Howe2011,Howe2013}; 2) reformulated features from existing binarization decision functions such those in \cite{Niblack1986} and \cite{Sauvola2000}; and 3) our newly developed features, namely the Logarithm Intensity Percentile (LIP) and the Relative Darkness Index (RDI). Our initial experimental results show that using only selected samples (about 1.5\% of all available training data), we can achieve a binarization performance comparable to those fine-tuned (typically by hand), state-of-the-art methods. Additionally, the trained document binarization classifier shows good generalization capabilities on out-of-domain data.
\end{abstract}

\IEEEpeerreviewmaketitle
\section{Introduction}
%
%
%
%
%
%
As one of the most fundamental preprocessing methods in various document analysis work \cite{DIBCO2009,DIBCO2010,DIBCO2011,DIBCO2012,DIBCO2013,DIBCO2014, Su2013, Combined2014, Howe2013,6065466}, document binarization aims to convert a color or grayscale document image into a monotonic image, where all text pixels of interest are marked in black with a white background. Mathematically, given a document image $\im{D}\!\!=\!\!\{D_{i,j}\}_{i\in[1,W], j\in[1,H]}$ of size $W\!\!\times\!\!H$, image binarization assigns each pixel $D_{i,j}$ a binary class label $B_{i,j}$ according to a decision function $f_{\rm{binarize}}(\cdot)$ in a meaningful way, namely
\begin{equation}
B_{i,j}=\left\{
\begin{array}{r@{}cl}
{\rm{foreground\,class\,}} 1 ,&&{\rm{if}}f_{\rm{binarize}}(D_{i,j}) < 0\\
{ \rm{background\,class\,}} 0 ,&&{\rm{else}} 
\end{array}
\right.
\label{eq.bin}
\end{equation}
A successful document binarization process discards irrelevant and noisy information while preserving meaningful information in the binary image $\im{B}\!\!=\!\!\{B_{i,j}\}$. This process reduces the space to represent a document image, and largely simplifies the complexity of advanced document analysis tasks \cite{Howe2011}.

Although human do not often face many difficulties in identifying texts even on some low-quality document images,  \dibp is indeed subjective and ill-posed \cite{Howe2011}, and it involves many different challenges and combinations of challenges. For example, several of the well-known ones are 1) how to handle document degradations like ink blob, fade text etc.; 2) how to deal with uneven lighting; and 3) how to differentiate bleed-through text from normal text. In such difficult scenarios, human actually uses high-level knowledge that might not be easily captured by low-level features--such as a script character set and background texture analysis--to help decide which pixel is foreground text. 

Classic solutions more or less seek heuristic thresholds in simple feature spaces. 
This can be further grouped into the so-called {\textit{global thresholding}} and {\textit{local thresholding}} methods \cite{Combined2014} according to whether this threshold is location independent or not.
For example, Otsu's method \cite{Otsu1975} binarizes a pixel $D_{i,j}$ by comparing its \textit{pixel intensity} $I_{i,j}$ to an optimal global threshold $G_{\rm{th}}$ derived from  \textit{intensity histogram} \cite{Otsu1975} as shown in \eqref{eq.otsu}
\begin{equation}
f_{\rm{Otsu}}(i,j)=I_{i,j} - G_{\rm{th}}
\label{eq.otsu}
\end{equation}
In contrast Niblack's method \cite{Niblack1986} uses the decision function \eqref{eq.niblack}
\begin{equation}
f_{\rm{Niblack}}(i,j)= I_{i,j}-\mu^R_{i,j} +k_{\rm{Niblack}}\sigma^R_{i,j} 
\label{eq.niblack}
\end{equation}
where $k_{\rm{niblack}}$ is a parameter below 0, and $\mu^R_{i,j}$ and $\sigma^R_{i,j}$ denote the mean and standard deviation of pixel intensities within a region $R$ of size $w\!\!\times\!\!h$. 
Although heuristic solutions are very efficient--may only requiring a constant number of operations per pixel, and work fairly well on many well-conditioned document images, it is clear that simple features and decision functions are insufficient for handling difficult cases. 

To achieve robust document binarization, many efforts are being made in the areas of 1) image normalization/adaptation, 2) discriminative feature space, and 3) more complicated decision functions. For example, Lu \etal \cite{Lu2010} proposes a local thresholding approach that mainly relies on background estimation and stroke estimation. Su \etal \cite{Su2009,Su2013} finds that Otsu's thresholding helps attain more discriminative power in a local contrast feature space. Sauvola \etal \cite{Sauvola1997,Sauvola2000} adds the parameter $S_{\rm{Sauvola}}$ to allow a non-linear decision plane \eqref{eq.sauvola}.
\begin{equation}
f_{\rm{Sauvola}}(i,j)\!\!=\!\! I_{i,j}-\mu^R_{i,j}\!-\!k_{\rm{Sauvola}}\mu^R_{i,j}(^{\sigma^R_{i,j}}/_{S_{\rm{Sauvola}}} \!\!-\!\!1 )  
\label{eq.sauvola}
\end{equation}
Although many of these attempts work well when method assumptions are satisfied and method parameters are appropriate, adapting a heuristic binarization method to a new domain is often not easy. Indeed, Lazzara et al. \cite{Sauvola2014} show that the original Sauovla method might fail even for well-scanned document images because of text fonts of different sizes. 

Unsupervised learning recently dominates document binarization area. In \cite{Su2012}, a document image is first clustered into three classes, namely foreground, background and uncertain, and pixels in the uncertain class will be further classified into either the foreground or background class according to their distances from these two classes. In \cite{Howe2011,Howe2013}, an image is first transformed into a Laplacian feature space, and a global energy function is constructed to  
ensure that resulting binary labels are optimal in the sense of a predefined Markov random field. In \cite{EoE2012}, an unsupervised ensemble of expert frameworks is used to combine multiple binarization candidates. Although these methods do not require a training stage, some rely on theoretical models or heuristic rules whose assumptions may not be necessarily satisfied, some require expensive iterative tuning and optimizations, and thus no surprise to see they are not reliable for certain types of degradations \cite{Phase2014}. 

Although image binarization is clearly a classification problem, supervised learning-based binarization solutions are still rare in the community. In this letter we discuss our initial attempts to solve the \dibp using supervised learning. The remainder of our paper is organized as follows: Section II overviews our solution and discusses all used features. Section III provides implementation details related to training and testing. Section IV shows our experimental results, and Section V concludes this paper.  

\section{Feature Engineering}
Our goal is to develop a generic solution without preset parameters and pre- or post-processing. Specifically, we are interested in learning a decision function $f_{\rm{ours}}(\cdot)$ that maps a $n$d feature vector $\vec{X}_{i,j}$ extracted around a pixel $D_{i,j}$ to a binary space $\{0,1\}$ in a meaningful way, i.e.
\begin{equation}
B_{i,j}=f_{\rm{ours}}(\vec{X}_{i,j})
\label{eq.binours}
\end{equation}
Detailed feature engineering discussions are given below.

\subsubsection{Existing Features}
Since a number of simple tasks can be accomplished just by applying Otsu's method. We thus include a pixel intensity $I_{i,j}$ and its deviation from the Otsu's threshold as features below
\begin{equation}
X_{i,j}^{\rm{Local int.}} = I_{i,j}
\end{equation}
\begin{equation}
X_{i,j}^{\rm{Otsu\,diff.}} = I_{i,j} - G_{\rm{th}}
\end{equation}
In addition, we also use local statistics of Eqs. \eqref{eq.localmean} and \eqref{eq.localdev}, but with respect to different scales, i.e.,
\begin{equation}\label{eq.localmean}
X_{i,j}^{\rm{Local\,avg.}|R}\!=\!\mu_{i,j}^{R}\!=\!\sum_{p\!=\!-^{w}\!/\!_{2}}^{^{w}\!/\!_{2}}\sum_{q\!=\!-^{h}\!/\!_{2}}^{^{h}\!/\!_{2}}{^{I_{i+p,j+q}}/_{wh}}
\end{equation}
\begin{equation}\label{eq.localdev}
X_{i,j}^{\rm{Local\,std.}|R} \!\!=\!\!\sigma_{i,j}^{R}\!=\!\sum_{p\!=\!-^{w}\!/\!_{2}}^{^{w}/_{2}}\sum_{q\!=\!-^{h}\!/\!_{2}}^{^{h}\!/\!_{2}}{^{I_{i+p,j+q}^2}/_{ wh}} \!-\!{\mu^R_{i,j}}^2
\end{equation}
where we make the size $w\!\!=\!\!h\!\!=\!\!ks$  of local window $R$ be associated with scales $k\in[1,2,4,8]$, and estimate stroke width $s$ using Su's method \cite{Su2013}. Inspired by the success of the Su \cite{Su2009,Su2013} and Howe methods \cite{Howe2011,Howe2013}, we include their contrast and Laplacian features shown in \eqref{eq.su} and \eqref{eq.howe}. 
\begin{equation}
X_{i,j}^{{\rm{Su}}|R}\!\!=\!\!{{\displaystyle\argmax_{p,q\in R}\{I_{i+p,j+q}\}-\displaystyle\argmin_{p,q\in R}\{I_{i+p,j+q}\}}\over{{\displaystyle\argmax_{p,q\in R}\{I_{i+p,j+q}\}\!+\!\displaystyle\argmin_{p,q\in R}\{I_{i+p,j+q}\}}}\!+\!\epsilon_{\rm{su}}}
\label{eq.su}
\end{equation}
\begin{equation}
X_{i,j}^{{\rm{Howe}}|R}=\nabla^2\mu_{i,j}^{R}
\label{eq.howe}
\end{equation}

\subsubsection{Exponential Truncated Niblack Index}
\begin{figure}[!h]
\centering\scriptsize
\includegraphics[width=\fwI\linewidth,frame]{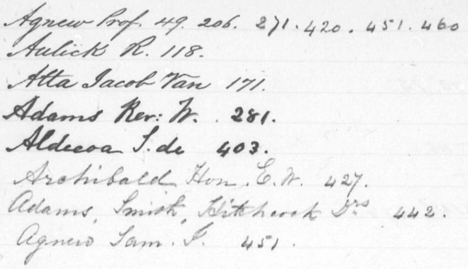}{(a)}
\includegraphics[width=\fwI\linewidth,frame]{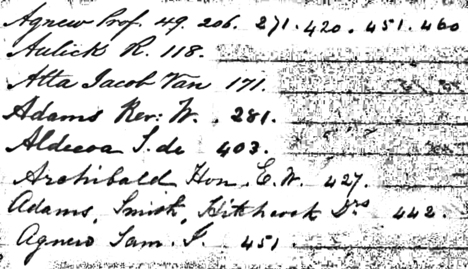}{(b)}
\caption{ETRI features for image DIBCO2010\_HW04. (a) Original image; (b) ETNI feature for $R$ of size $64\times 64$. }\label{fig.etni}
\end{figure}

To include Niblack's decision function in our considerations, we first rearrange terms in \eqref{eq.niblack} according to $f_{\rm niblack}\!\!=\!\!0$, as shown below
\begin{equation}
k_{\rm{Niblack}}(i,j|R)= (I_{i,j}-\mu_{i,j}^R)/\sigma_{i,j}^{R}
\end{equation} 
and then compute a so-called Exponential Truncated Niblack Index (ETNI) feature as follows.
\begin{equation}
X_{i,j}^{{\rm{ETNI}}|R} =\left\{
\begin{array}{rl}
\exp\{k_{\rm{Niblack}}(i,j|R)\},&{\rm if} I_{i,j}\leq\mu_{i,j}^R\\
1,&\rm otherwise
\end{array}
\right.
\label{eq.TNI}
\end{equation}
Fig. \ref{fig.etni} compares an image in the original form and its corresponding ETRI feature space. 

\subsubsection{ Logistic Truncated Sauvola Index}
\begin{figure}[!h]
\centering\scriptsize
\includegraphics[width=\fwI\linewidth,frame]{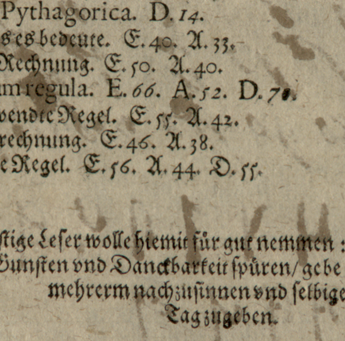}{(a)}
\includegraphics[width=\fwI\linewidth,frame]{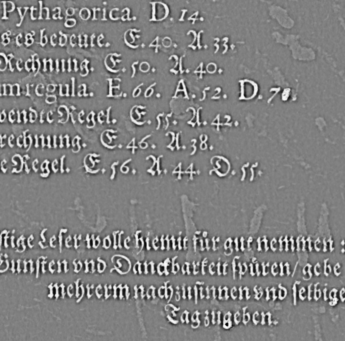}{(b)}
\caption{LTSI features for image DIBCO2011\_PR05. (a) Original image; (b) LTSI feature for $R$ of size $8\times 8$. }\label{fig.ltsi}
\end{figure}

Similarly, we rearrange terms in Sauvola's decision function \eqref{eq.sauvola} according to $f_{\textrm{Sauvola}}=0$ for its key parameter $k_{\rm{Sauvola}}$ as follows,
\begin{equation}
k_{\rm{Sauvola}}(i,j|R) \propto \displaystyle{I_{i,j}/\mu_{i,j}^R-1\over \sigma_{i,j}^{R}-S_{\rm{Sauvola}}}
\end{equation} 
Since $k_{\rm{Sauvola}}(i,j|R)$ could be $(-\infty,\infty)$, we normalize this index by using the logistic function shown in Eq. \eqref{eq.TSI}, and call it the Logistic Truncated Sauvola Index (LTSI),
\begin{equation}
X_{i,j}^{{\rm{LTSI}}|\!R} \!\!\!\!=\!\!
\left\{\!\!\!\!
\begin{array}{c@{}l}
0&,{\rm if} \sigma_{i,j}^{R}\!\!>\!\!S_{\rm{Sauvola}}\\
( \!1\!\!+\!\!\exp\!\{\!-\!\!k_{\rm{Sauvola}}(\!i,\!j|R)\}\!)\!^{\tiny{-\!1}}&,\rm otherwise
\end{array}
\right.
\label{eq.TSI}
\end{equation}
where the range of $X_{i,j}^{{\rm{TSI}}|R}$ is $[0\!,\!1]$, and the condition $\sigma_{i,j}^{R}\!<\!S_{\rm{Sauvola}}$ ensures the sign consistency of $k_{\rm{Sauvola}}(i,j|R)$. LTSI thus reflects the Sauvola decision surface. A sample result of the LTSI feature is given in Fig. \ref{fig.ltsi}.

\subsubsection{Logarithm Intensity Percentile Features}
Intuitively, the darkness of a pixel is related to whether it is a text pixel. Given a region $S$, the percentile of the pixel's intensity can be computed as
\begin{equation}
{\rm perc}(i,j|S) = \sum_{p,q\in S} {\im{1}_{[0,\infty)}(I_{i,j}- I_{i+p,j+q})\over \|S\|}
\label{eq.perc}
\end{equation}
where $\im{1}_{[0,\!\infty)}(t)$ denotes the indicator function whose value is 1 when $t\!\in\![0\!,\!\infty)$ and 0 otherwise, and $\|\!\cdot\!\|$ denotes the cardinality function. It is clear that this percentile is a type of rank feature, and thus is invariant to any monotonic transform on the original intensity space. To give a higher resolution for lower percentiles, we use the logarithm version of \eqref{eq.perc} as shown in \eqref{eq.LIP}, and call it Logarithm Intensity Percentile (LIP) feature. Here $\rm{Th}_{perc}$ is a threshold ( =.01 in this paper). 
\begin{equation}
X_{i,j}^{{{\rm{LIP}}|S}} =\left\{
\begin{array}{l}
\!\!1.0, {\rm if}\,{\rm perc}(i,j|S)\leq\rm{{Th}_{perc}}\\
\!\!\displaystyle\log_{\rm{{Th}_{perc}}}({\rm perc}(i,j|S) ), {\rm otherwise}
\end{array}
\right.
\label{eq.LIP}
\end{equation}

\begin{figure}[!h]
\centering\scriptsize
\begin{tabular}{@{}c@{}m{.05cm}@{}c@{}m{.05cm}@{}c@{}m{.05cm}@{}c@{}m{.05cm}@{}c@{}m{.05cm}@{}c@{}}
\includegraphics[width=\fwIb\linewidth,frame]{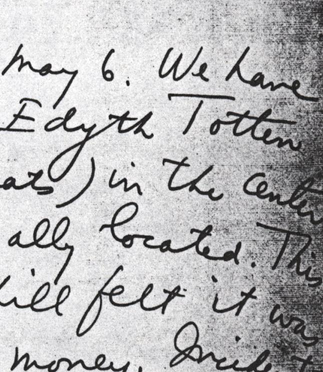}&&
\includegraphics[width=\fwIb\linewidth,frame]{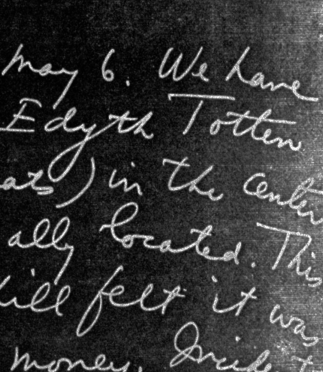}&&
\includegraphics[width=\fwIb\linewidth,frame]{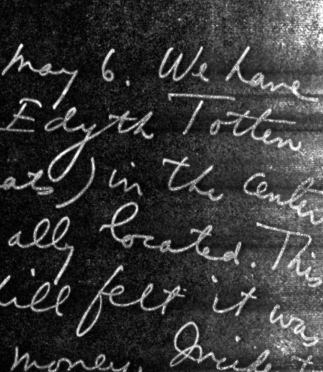}&&
\includegraphics[width=\fwIb\linewidth,frame]{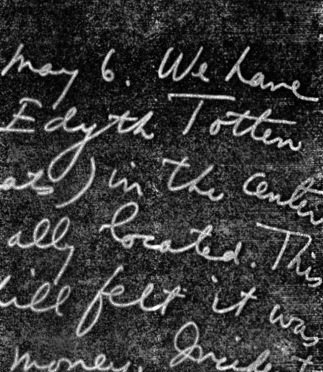}&&
\includegraphics[width=\fwIb\linewidth,frame]{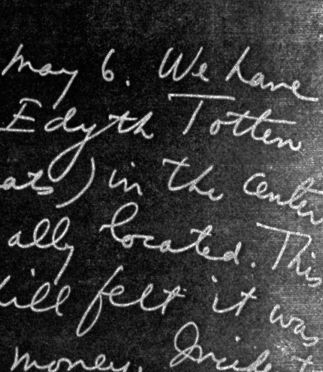}&&
\includegraphics[width=\fwIb\linewidth,frame]{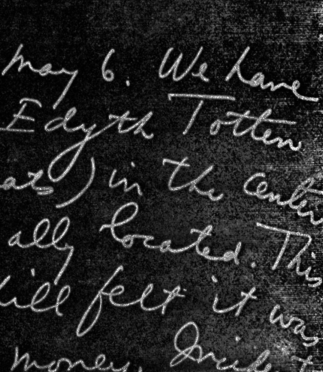}\\
{(a)}&&
{(b)}&&
{(c)}&&
{(d)}&&
{(e)}&&
{(f)}\\
\end{tabular}
\caption{LIP features for image DIBCO2011\_HW1. (a) original  image; (b) global LIP; (c)-(e) LIP along row, column, and diagonal; and (f) max LIP of all directions.}\label{fig.lip}
\end{figure}

With regard to $S$, we make parallelogram $S$ cover multiple rows, columns, diagonals, and inverse diagonals. The number of rows, columns, diagonals and inverse diagonals in $S$ is made to be $k$ times the estimated stroke width $s$. Finally, we also compute the LIP features with respect to the entire image and the maximum percentile among all previously extracted LIP features. Fig. \ref{fig.lip} shows the original document with its corresponding features in the LIP spaces. As one can see, the LIP space indeed provides more discriminative powers. 

\subsubsection{Relative Darkness Index Features}
Inspired by the great success of local ternary patterns(LTP) \cite{LTP2010} in face recognition, we borrow their essences here. LTP relies on the comparison of a center pixel's intensity with each pixel in a set of neighbors $\{N_1, \cdots, N_k\}$ that are on a radius $r$ circle, and the $l$th code in a length-$k$ code string is defined as
\begin{equation}
{\rm{ltp}}( P_{i,j}, l ) = \left\{
\begin{array}{rcl}
+1, &{\rm if}& I_{i +r_l, j+c_l} \geq I_{i,j} + \rm{tol}\\
-1, &{\rm if}& I_{i +r_l, j+c_l} \leq I_{i,j}-\rm{tol}\\
0, &{\rm if }& | I_{i +r_l, j+c_l} - I_{i,j}|<\rm{tol}\\
\end{array}
\right.
\end{equation}
where $r_l$ and $c_l$ denote the relative coordinates of a neighbor $N_l$ w.r.t. a center pixel, and $\rm{tol}$ is a preset tolerance. However, the number of possible LTP codes is often huge to effectively encode. Though one may reduce this number by considering all shift-equivalent codes as one, or separating a ternary code into two binary codes, we find that the simple frequency count of each code in a code string has already revealed many intrinsic properties of pixels, and we call them the Relative Darkness Index (RDI) features. 
Precisely, given the code ${\cal{C}}$ and neighbors on a radius $r$ circle, the RDI feature can be defined as below
\begin{equation}
X_{i,j}^{{\rm{RDI}}| {\cal{C}}, r } = \sum_{l=1}^{k}{{\im{1}}_{0}({\rm{ltp}}(P_{i,j}, l)-{{\cal{C}}})\over k}.
\label{fig.rdi}
\end{equation}
As one can see from Fig.\ref{fig.rdi}(c-e), most of the nearly homogeneous background parts are of high code 0 indices; pixels close to strong edges are dominated by code+1 indices, and foreground text pixels have high response on code-1 indices. To further enhance RDI's discriminative power, we compute the ratios of one code to the sum of itself and another code as well (see Fig.\ref{fig.rdi}(f-h)).
\begin{figure}[!ht]
\centering\scriptsize
\includegraphics[width=\fwII\linewidth,frame]{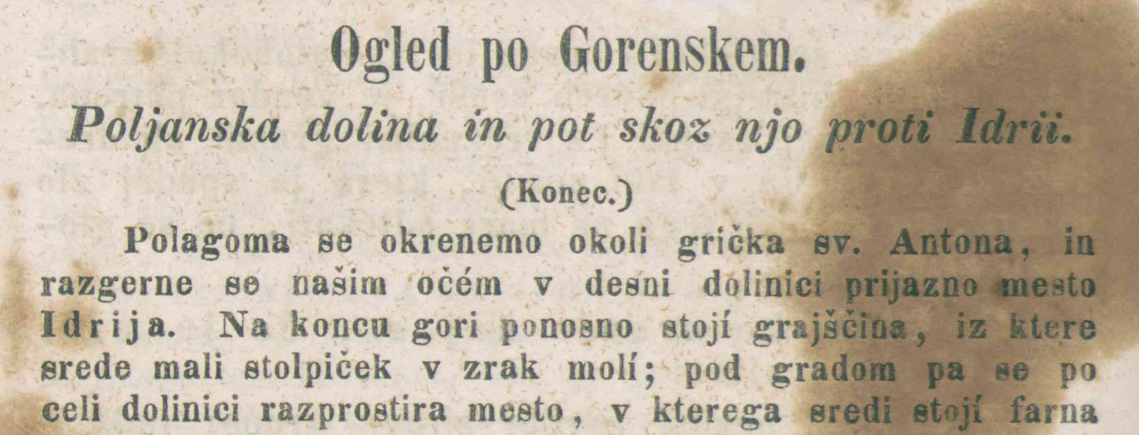}{(a)}
\includegraphics[width=\fwII\linewidth,frame]{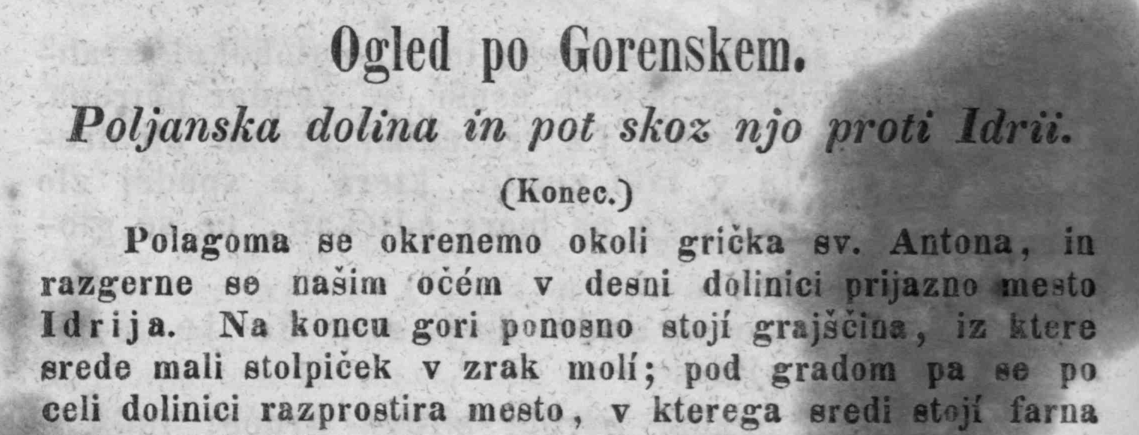}{(a)}
\includegraphics[width=\fwII\linewidth,frame]{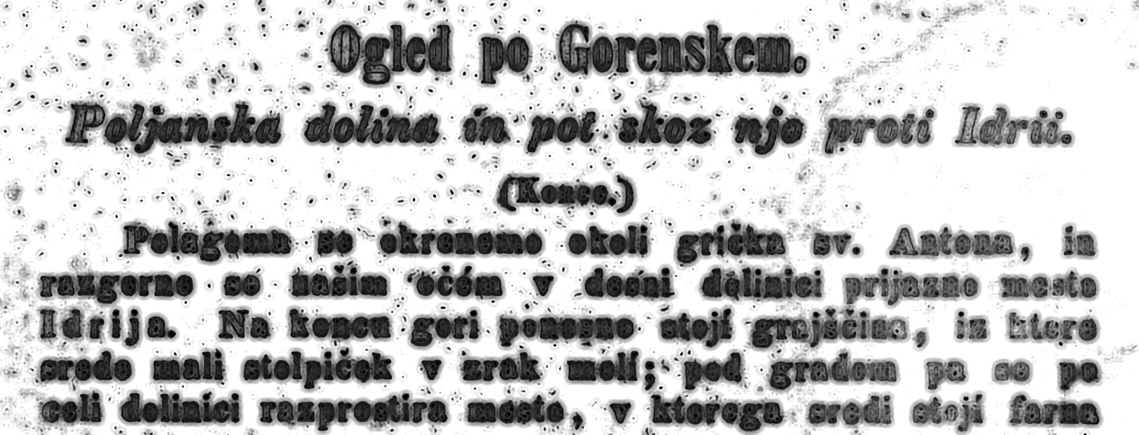}{(c)}
\includegraphics[width=\fwII\linewidth,frame]{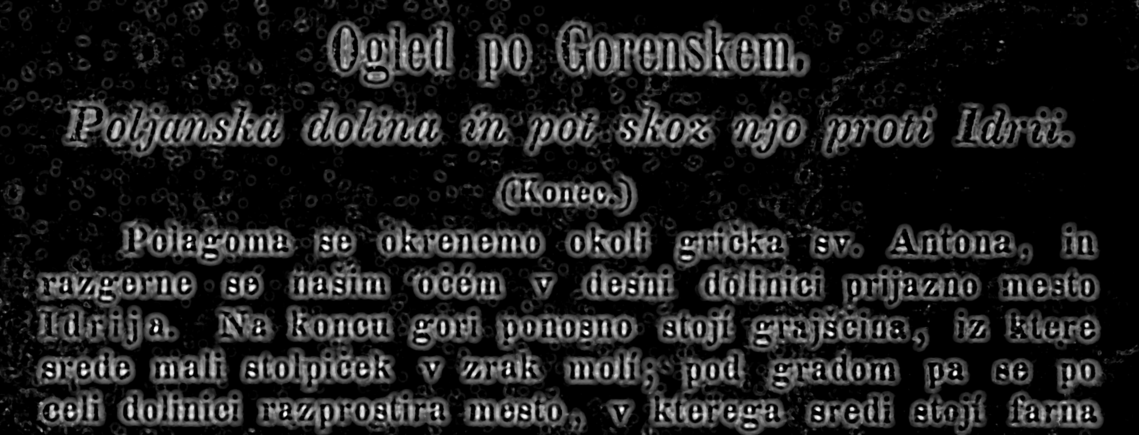}{(d)}
\includegraphics[width=\fwII\linewidth,frame]{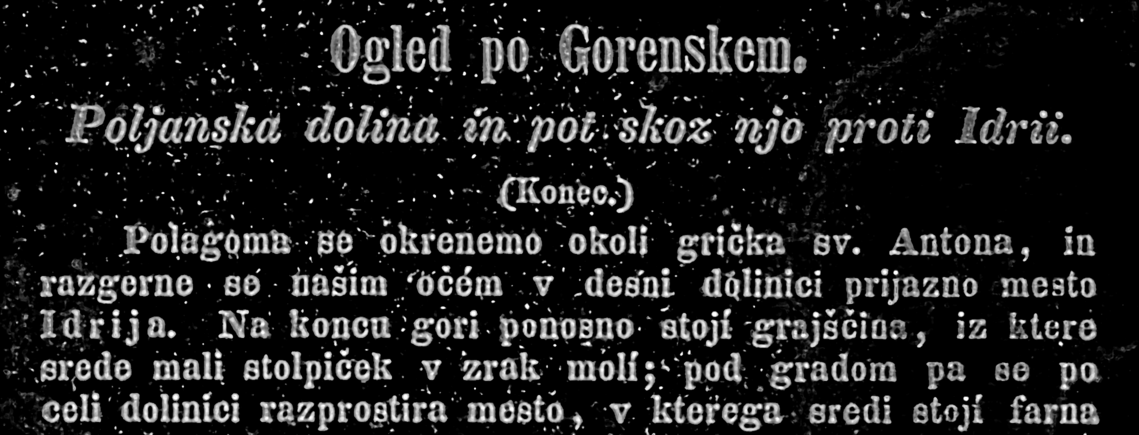}{(e)}
\includegraphics[width=\fwII\linewidth,frame]{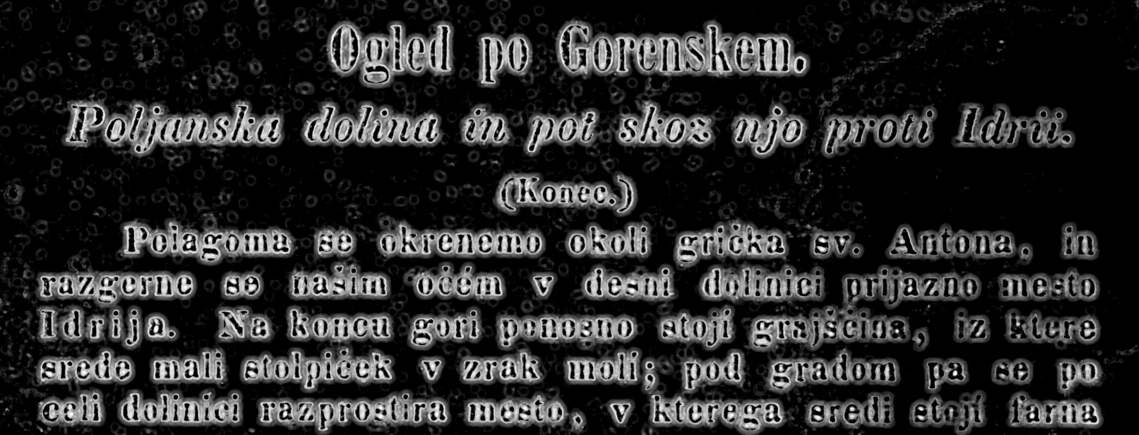}{(f)}
\includegraphics[width=\fwII\linewidth,frame]{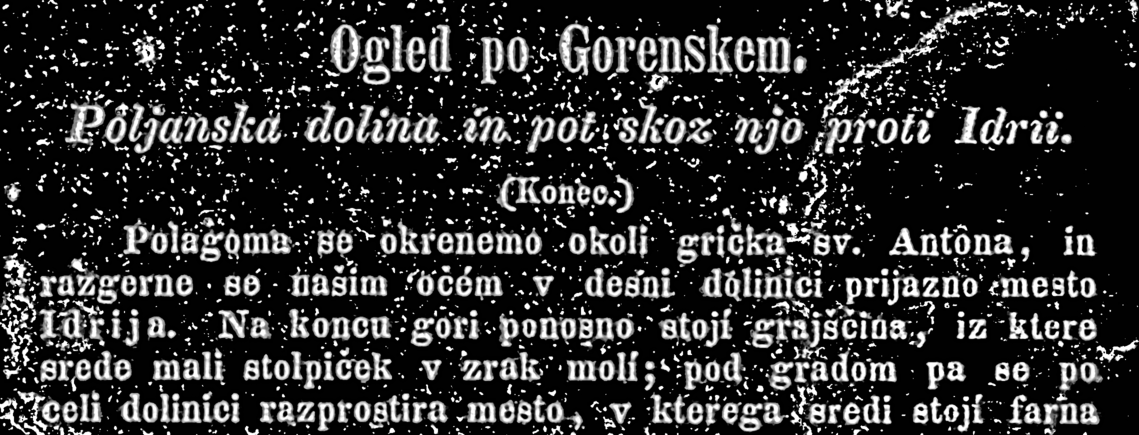}{(g)}
\includegraphics[width=\fwII\linewidth,frame]{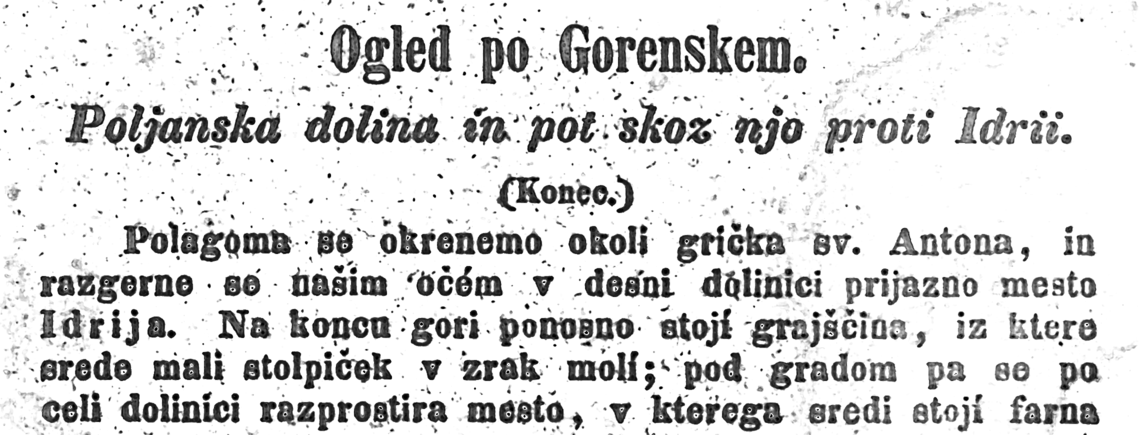}{(h)}
\caption{RDI features for image DIBCO2013\_PR05 (darker pixels indicate a value close to 0). (a) original color image; (b) original image; (c)-(e) RDI feature $X^{{\rm{RDI}}| {\cal{C}}, 8 }$for ${\cal{C}}\in\{0\!,\!-1\!,\!+1\}$, respectively; (f) ${ X^{{\rm{RDI}}| {\cal{C}}\!=\!+\!1, 8 }\over X^{{\rm{RDI}}| {\cal{C}}\!\in\!\{0,+1\}, 8 }}$ ; (g)${ X^{{\rm{RDI}}| {\cal{C}}\!=\!-\!1, 8 }\over X^{{\rm{RDI}}| {\cal{C}}\!\in\!\{-1,+1\}, 8 }}$ ; and (h) ${ X^{{\rm{RDI}}| {\cal{C}}\!=\!+\!0, 8 }\over X^{{\rm{RDI}}| {\cal{C}}\!\in\!\{-1,0\}, 8 }}$ .}
\label{fig.rdi}
\end{figure}
\subsubsection{Other Features}
Besides of features discussed above, we extract features from the global image statistics, including the mean and standard deviation of the entire image intensities, the mean and standard deviation of the percentile image, the 32 bins of normalized histogram (sum to 1) for image intensities, and the 32 bins of a normalized logarithmed histogram. 

\section{Training and Testing Settings}
\setcounter{subsubsection}{0}
In experiments, we use the widely accepted Document Image Binarization Contest (DIBCO) from 2009 to 2014 \citeDIBCO as our training and testing data; it totals 76 images. We adopt the leave-one-out strategy where we first pick a DIBCO image set of a particular year as our testing set, and use the rest as our training set.  

\subsubsection{Feature Summary}
We summarize all used features with dimensions and corresponding normalization considerations in Table \ref{tab.feat}. Here, the stroke width $s$ can be estimated via various methods; we use Su's method \cite{Su2013}. `Scale' indicates the side of local square region $R$. 
\begin{table}[!h]
\centering
\scriptsize
\caption{Used Features }\label{tab.feat}
\begin{tabular}{rrrr}
\hline
\bf{Type}&\bf{Scale}&\bf{Dimension}&\bf{Normalization}\\\hline
Local int.&N/a&1&divide by 255\\
Otsu diff.&N/a&1&divide by 255\\
Local avg./std.&1$s$,2$s$,4$s$,8$s$&4/4&divide by 255\\
Su/Howe&1,1$s$,2$s$,4$s$&4/4&MinMax\\
ETRI/LTSI&1$s$,2$s$,4$s$,8$s$&4/4&N/a\\
LIP&1,1$s$,2$s$,4$s$,8$s$&1+4$\times$4+1&N/a\\
RDI&1,1$s$,2$s$,4$s$,8$s$&5$\times$6&N/a\\
Global int. avg./std.&N/a&1/1&divide by 255\\
Global perc. avg./std.&N/a&1/1&N/a\\
Global int./perc. loghist. &N/a&32/32&N/a\\\hline
Total& & 142 &\\
\hline
\end{tabular}
\end{table}

\subsubsection{Sampling Strategy}
Selecting training samples is essential in task. First, one may not be handle a big training set of this task. These 76 images totally contain more than 80 million pixels. Assuming each feature is store in {\sf{float32}} format, we need $80\!\!\times \!\!142\!\!\times\!\!4$MB ($\gg$256GB) memory for just training features, while this requirement clearly beyond the capacities of most computers nowadays. Second, one may notice the imbalanced training data. We know both the background nontext class and foreground text class in the binarization problem actually cover different subclasses \cite{Su2012,ramirez2010transition}, while we also know nearly homogeneous background and foreground dominate our training data. 

\begin{figure}[!h]
\centering
\scriptsize
\includegraphics[width=.33\linewidth,frame]{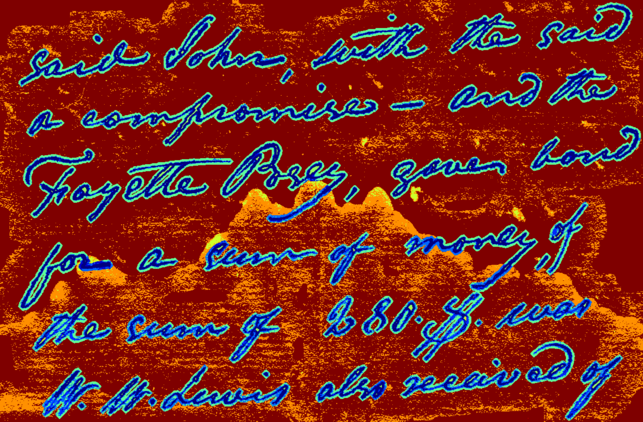}{(a)}
\includegraphics[width=.33\linewidth,frame]{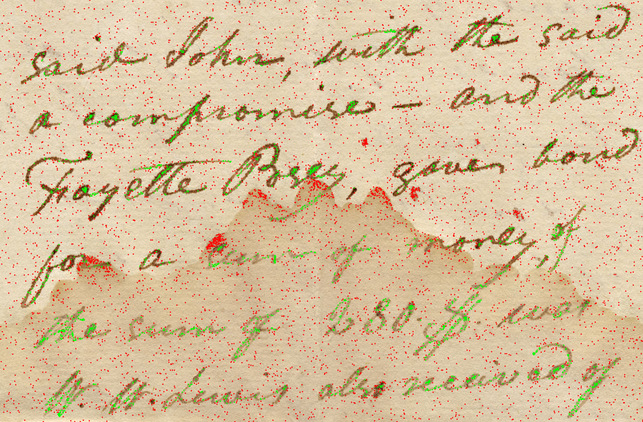}{(b)}
\caption{Sampling strategy. (a) pixels with subclass labels for image DIBCO2012-HW02 (each color denotes a subclass); (b) samples extracted from DIBCO2012-HW02 image balanced subclasses (red/green dots indicate background/foreground.)}
\label{fig.sampling}
\end{figure}

To solve both problems, we first artificially classify all pixels in an image into 16 subclasses, each is represented as a 4-bit string, where each bit indicates whether or not this pixel should be treated as a pixel in Otsu's foreground, in Niblack's foreground, within $s$ pixels away from reference image edges, and in a reference annotated foreground. We draw the same number of random samples for each subclass. Fig. \ref{fig.sampling} illustrates the samples we extracted that balanced both foreground and background subclasses. 

\subsubsection{Training and Testing Strategies}
\begin{figure}[!h]
\centering
\scriptsize
\includegraphics[width=0.495\linewidth,trim= 4.5cm 0.5cm 1.5cm 2cm,clip=true]{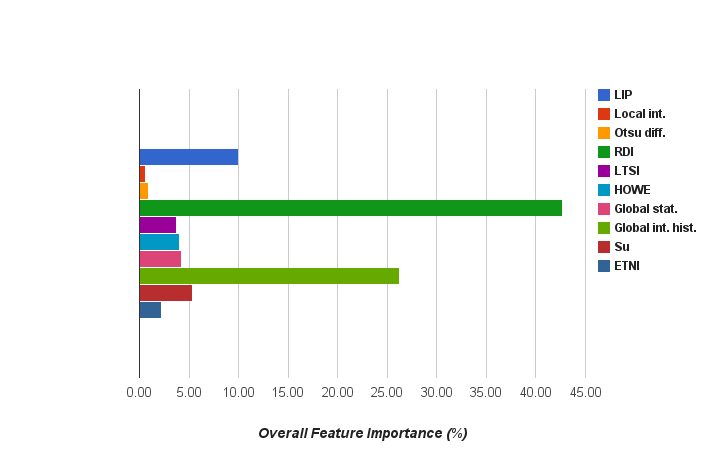}
\includegraphics[width=0.495\linewidth,trim= 4.5cm 0.5cm 1.5cm 2cm,clip=true]{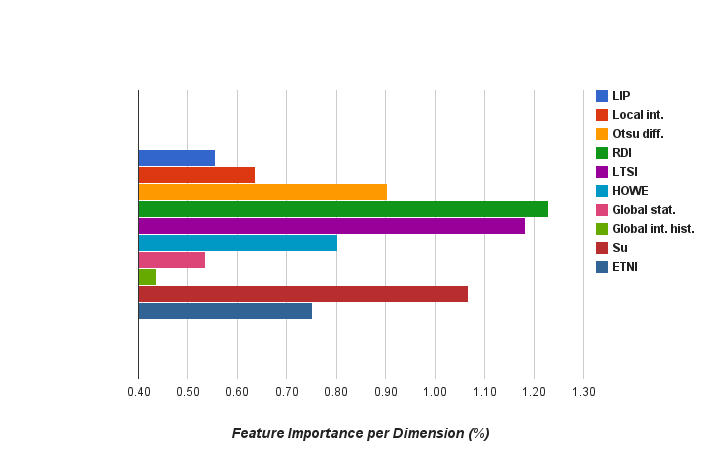}
\caption{Feature importance. Left: overall importance of each feature type; and Right: dimensional feature importance for each feature type. }
\label{fig.importance}
\end{figure}

In all of the following experiments, we perform a two-pass training. We first extract 9,600 samples (subclass balanced) from each training image and train a simple classifier, say Gaussian Naive Bayes. We use this classifier to decode all training images, and extract additional 9,600 erroneous samples (subclass balanced) from each image, and use  all extracted samples to train an more complicated {\textsl{sklearn}} \cite{sklearn} {\textit{ ExtraTrees}} classifier \cite{geurts2006extremely}. Note in total we extract 19,200 samples per image, which only account for roughly about 1.5\% of all samples. Classifier parameters are obtained from a 10-folded cross-validation using all samples. A final classifier is trained by using all extracted samples and validated parameters. Fig. \ref{fig.importance} plots the feature importance of each feature type in terms of the overall contribution and the averaged dimensional contribution with respect to each feature type. As one can see, RDI, Global int. hist. and LIP are the three most useful feature categories in terms of overall contributions; and Su, LTSI and RDI are the three best features in terms of dimensional contributions. 

In testing, we use the final classifier to predict the class label for all pixels in a testing image. Depending on the size of an image, the decoding time may vary between 5s to 30s.

\section{Experimental Results}\setcounter{subsubsection}{0}
\subsubsection{Performance on DIBCO Datasets}
\begin{figure}[!h]
\centering\scriptsize
\begin{tabular}{@{}c@{}m{.05cm}@{}c@{}}
{DIBCO2012\_H07}&&{DIBCO2014\_HW07}\\
\includegraphics[height=\fhIb,frame]{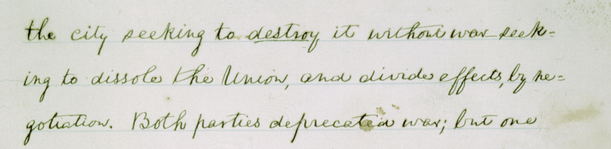}&&\includegraphics[height=\fhIb,frame]{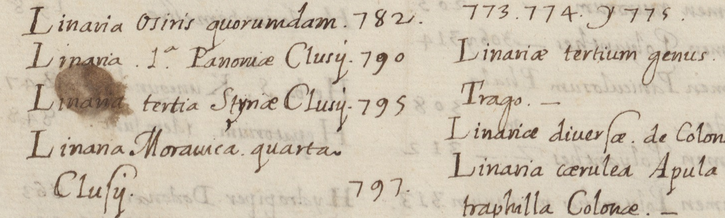}{(a)}\\
\includegraphics[height=\fhIb,frame]{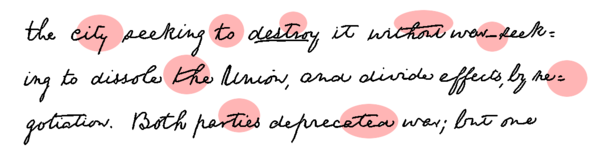}&&\includegraphics[height=\fhIb,frame]{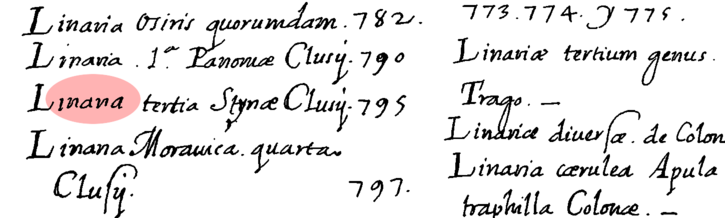}{(b)}\\
\includegraphics[height=\fhIb,frame]{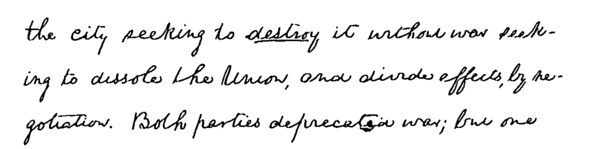}&&\includegraphics[height=\fhIb,frame]{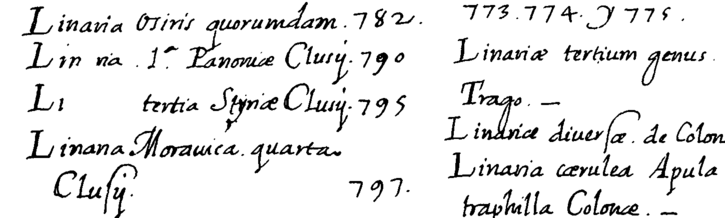}{(c)}\\
\includegraphics[height=\fhIb,frame]{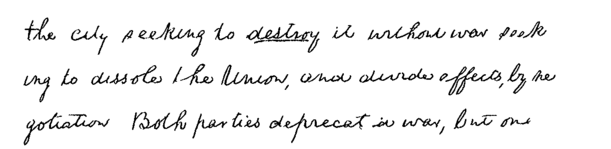}&&\includegraphics[height=\fhIb,frame]{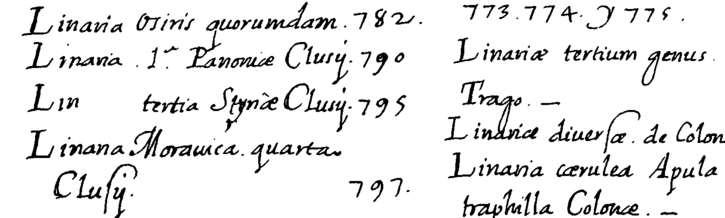}{(d)}\\
\includegraphics[height=\fhIb,frame]{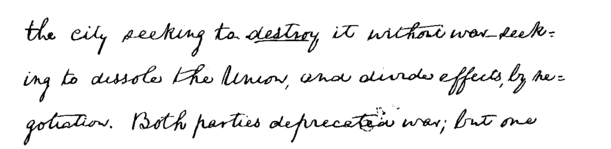}&&\includegraphics[height=\fhIb,frame]{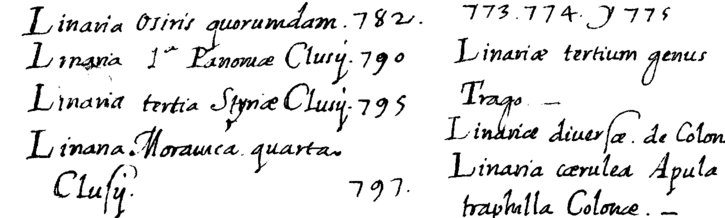}{(e)}
\end{tabular}
\caption{Binarization results for image. (a) original images; (b) reference binarized images (highlighted red regions indicate disagreements); (c) results of contest rank 1; (d) results of contest rank 2; and (e) our results.}
\label{fig.2014}
\end{figure}

Table II lists performance of our proposed supervised binarization solution over the DIBCO 2012 \cite{DIBCO2012}, 2013 \cite{DIBCO2013}, and 2014 \cite{DIBCO2014} datasets using standard metrics F1-score, peak signal-to-noise ratio (PSNR), and distance reciprocal distortion (DRD) ( metric definitions can be found in  \citeDIBCO). As we can see, our performance is comparable to the top five methods. We also notice that our binarization classifier's performance is very stable among all three datasets, especially since it always keeps a DRD score below 3 pixels. Sample decoding results are compared to the top two contest methods in Fig. \ref{fig.2014}. As one can see, our supervised solution successfully learnt knowledge to handle difficult cases: 1) faded text; and 2) text on a dirty background.  
\begin{table}[!h]
\centering\scriptsize
\caption{ Performance Evaluations On DIBCO Datasets}
\begin{tabular}{l|lrrrr}
\hline
&\bf{Method} & \bf{Contest Rank} & \bf{F1}\% & \bf{PSNR} & \bf{DRD} \\\hline
\multirow{6}{*}{\begin{sideways}{\bf{DIBCO2012}}\end{sideways}}&\cite{Howe2011} & 1 & 89.47 & 21.80 & 3.400\\
&Lelore \etal's \cite{DIBCO2012}   & 2 & 92.85 & 20.57 & 2.660\\
&\cite{Su2009}  & 3 & 91.54 & 20.14 & 3.048\\
&Nina's \cite{DIBCO2012}    & 4 & 90.38 & 19.30 & 3.348\\
&Yazid \etal's  \cite{DIBCO2012} & 5 & 91.85 & 19.65 & 3.056\\\cline{2-6}
&Ours   &   & 92.01 & 19.92 & 2.601\\
\hline
\multirow{6}{*}{\begin{sideways}{\bf{DIBCO-2013}}\end{sideways}}&Su \etal 's method \cite{DIBCO2013} & 1 & 92.12 & 20.68 & 3.100 \\
&\cite{Howe2013}  & 2 & 92.70 & 21.29 & 3.180 \\
&\cite{EoE2012}  & 3 & 91.81 & 20.68 & 4.020 \\
&\cite{6482566} & 4 & 91.69 & 20.54 & 3.590 \\
&\cite{ramirez2010transition}   & 5 & 90.92 & 19.32 & 3.910 \\\cline{2-6}
&Ours   &   & 91.40 & 20.13 & 2.637\\
\hline
\multirow{6}{*}{\begin{sideways}{\bf{DIBCO-2014}}\end{sideways}}&Mesquita \etal 's \cite{DIBCO2014} & 1 & 96.88 & 22.66 & 0.902 \\
&\cite{Howe2013}  & 2 & 96.63 & 22.40 & 1.001 \\
&\cite{nafchi2013historical}  & 3 & 93.35 & 19.45 & 2.194 \\
&Ziaratban \etal 's \cite{DIBCO2014}   & 4 & 89.24 & 18.94 & 4.502 \\
&Mitianoudis \etal 's \cite{DIBCO2014}      & 5 & 89.77 & 18.49 & 4.502 \\\cline{2-6}
&Ours   &   & 92.69 & 19.47 & 2.571\\
\hline
\end{tabular}
\end{table}

\subsubsection{Learning Curve}
As we mentioned previously, only about 1.5\% of all available training samples are used in our experiments. We investigate the relationship between the amount of training samples and the binarization performance using the test set of DIBCO 2012 in Table \ref{tab.sample}.  As in many pattern recognition problems, the improvement of binarization performance gets smaller as the number of samples increases. 

\begin{table}[!h]
\centering\scriptsize
\caption{ Performance v.s. Training Samples }
\label{tab.sample}
\begin{tabular}{r|rrrrrrr}
\hline
\bf{\#Samples} &\bf{1,920} &\bf{5,760} &\bf{9,600} &\bf{13,440} &\bf{15,360} &\bf{17,280} &\bf{19,200} \\\hline
 \bf{F1}\% & 91.47 & 91.77 & 91.90 & 91.93 & 91.95 & 92.01 & 92.01 \\
 \bf{PSNR} & 19.64 & 19.81 & 19.86 & 19.86 & 19.88 & 19.93 & 19.92 \\
 \bf{DRD} & 2.797 & 2.689 & 2.637 & 2.634 & 2.618 & 2.599 & 2.601 \\
 \hline
\end{tabular}
\end{table}

\subsubsection{Document Binarization in the Wild }
\begin{figure}[!h]
\centering
\begin{tabular}{c@{}c@{}c@{}}
\includegraphics[height=\fhI,frame]{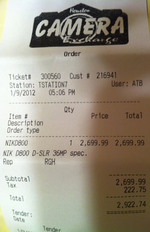}&\includegraphics[height=\fhI,frame]{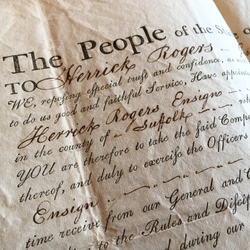}&\includegraphics[height=\fhI,frame]{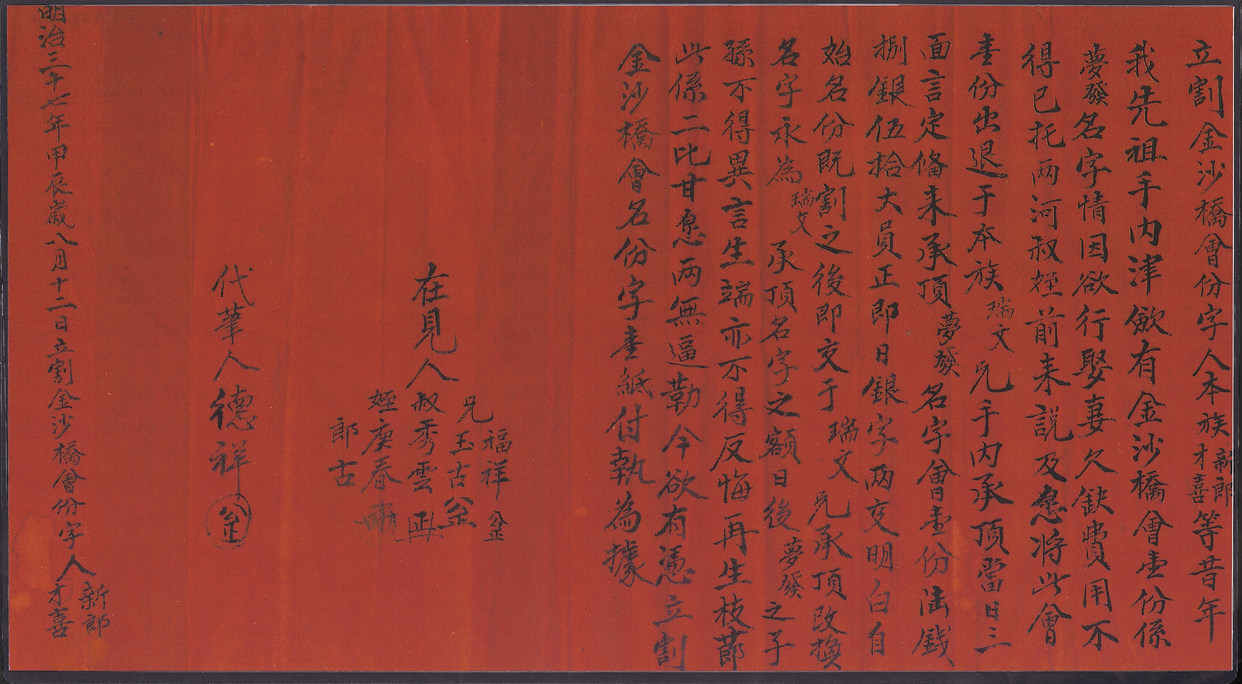}\\
\includegraphics[height=\fhI,frame]{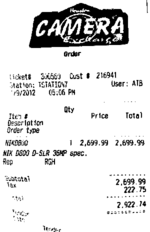}&\includegraphics[height=\fhI,frame]{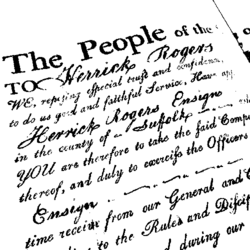}&\includegraphics[height=\fhI,frame]{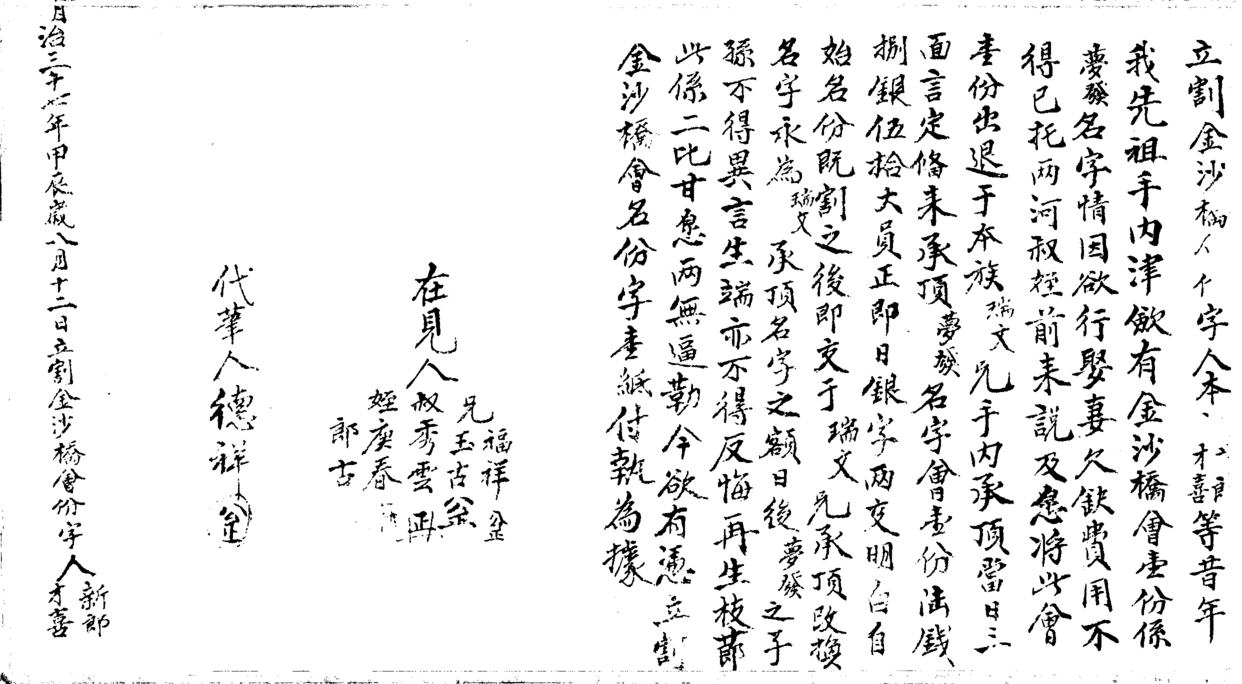}\\
\end{tabular}
\caption{Binarization results of out-of-domain data}
\label{fig.odd}
\end{figure}

Although images in DIBCO datasets have already covered a wide range of variations, there are clearly more variations and combinations of variations that are not included in DIBCO training data. We therefore test our learned classifier on out-of-domain document images, and we observe satisfactory results (see Fig. \ref{fig.odd}).

%
%

\section{Conclusion}
In this paper we investigate the document binarization solution via supervised learning. Unlike previous efforts, this solution is parameter-free and fully trainable. Our experimental results showed that one can learn a reasonably well binarization decision function from a small set of carefully selected training data. Such a learned decision function not only works well for in-domain data, but can also apply to out-of-domain data.
In future work, we will explore several interesting aspects such as discriminative features (e.g., image moments and connected component attributes) and classifier adaptation on the fly.

\ifCLASSOPTIONcaptionsoff
  \newpage
\fi
\newpage
\IEEEtriggeratref{8}
\IEEEtriggercmd{\enlargethispage{-5in}}


\begin{thebibliography}{10}
\providecommand{\url}[1]{#1}
\csname url@samestyle\endcsname
\providecommand{\newblock}{\relax}
\providecommand{\bibinfo}[2]{#2}
\providecommand{\BIBentrySTDinterwordspacing}{\spaceskip=0pt\relax}
\providecommand{\BIBentryALTinterwordstretchfactor}{4}
\providecommand{\BIBentryALTinterwordspacing}{\spaceskip=\fontdimen2\font plus
\BIBentryALTinterwordstretchfactor\fontdimen3\font minus
  \fontdimen4\font\relax}
\providecommand{\BIBforeignlanguage}[2]{{%
\expandafter\ifx\csname l@#1\endcsname\relax
\typeout{** WARNING: IEEEtran.bst: No hyphenation pattern has been}%
\typeout{** loaded for the language `#1'. Using the pattern for}%
\typeout{** the default language instead.}%
\else
\language=\csname l@#1\endcsname
\fi
#2}}
\providecommand{\BIBdecl}{\relax}
\BIBdecl

\bibitem{Otsu1975}
N.~Otsu, ``A threshold selection method from gray-level histograms,''
  \emph{Automatica}, vol.~11, no. 285-296, pp. 23--27, 1975.

\bibitem{Su2009}
B.~Su, S.~Lu, and C.~L. Tan, ``Binarization of historical document images using
  the local maximum and minimum,'' in \emph{Proceedings of the 9th IAPR
  International Workshop on Document Analysis Systems}.\hskip 1em plus 0.5em
  minus 0.4em\relax ACM, 2010, pp. 159--166.

\bibitem{Su2013}
------, ``Robust document image binarization technique for degraded document
  images,'' \emph{Image Processing, IEEE Transactions on}, vol.~22, no.~4, pp.
  1408--1417, 2013.

\bibitem{Howe2011}
N.~R. Howe, ``A laplacian energy for document binarization,'' in \emph{Document
  Analysis and Recognition (ICDAR), 2011 International Conference on}.\hskip
  1em plus 0.5em minus 0.4em\relax IEEE, 2011, pp. 6--10.

\bibitem{Howe2013}
------, ``Document binarization with automatic parameter tuning,''
  \emph{International Journal on Document Analysis and Recognition (IJDAR)},
  vol.~16, no.~3, pp. 247--258, 2013.

\bibitem{Niblack1986}
W.~Niblack, \emph{An Introduction to Digital Image Processing}.\hskip 1em plus
  0.5em minus 0.4em\relax Prentice-Hall, 1986.

\bibitem{Sauvola2000}
J.~Sauvola and M.~Pietik{\"a}inen, ``Adaptive document image binarization,''
  \emph{Pattern recognition}, vol.~33, no.~2, pp. 225--236, 2000.

\bibitem{DIBCO2009}
B.~Gatos, K.~Ntirogiannis, and I.~Pratikakis, ``Icdar 2009 document image
  binarization contest (dibco 2009),'' in \emph{Document Analysis and
  Recognition (ICDAR), 2009 International Conference on}, vol.~9, 2009, pp.
  1375--1382.

\bibitem{DIBCO2010}
I.~Pratikakis, B.~Gatos, and K.~Ntirogiannis, ``H-dibco 2010-handwritten
  document image binarization competition,'' in \emph{Frontiers in Handwriting
  Recognition (ICFHR), 2010 International Conference on}.\hskip 1em plus 0.5em
  minus 0.4em\relax IEEE, 2010, pp. 727--732.

\bibitem{DIBCO2011}
------, ``Icdar 2011 document image binarization contest (dibco 2011),'' in
  \emph{Document Analysis and Recognition (ICDAR), 2011 International
  Conference on}, 2011, pp. 1506--1510.

\bibitem{DIBCO2012}
------, ``Icfhr 2012 competition on handwritten document image binarization
  (h-dibco 2012).'' \emph{ICFHR}, vol.~12, pp. 18--20, 2012.

\bibitem{DIBCO2013}
------, ``Icdar 2013 document image binarization contest (dibco 2013),'' in
  \emph{Document Analysis and Recognition (ICDAR), 2013 International
  Conference on}.\hskip 1em plus 0.5em minus 0.4em\relax IEEE, 2013, pp.
  1471--1476.

\bibitem{DIBCO2014}
K.~Ntirogiannis, B.~Gatos, and I.~Pratikakis, ``Icfhr2014 competition on
  handwritten document image binarization (h-dibco 2014),'' in \emph{2014 14th
  International conference on frontiers in handwriting recognition}, 2014, pp.
  809--813.

\bibitem{Combined2014}
------, ``A combined approach for the binarization of handwritten document
  images,'' \emph{Pattern Recognition Letters}, vol.~35, pp. 3--15, 2014.

\bibitem{6065466}
X.~Peng, H.~Cao, R.~Prasad, and P.~Natarajan, ``Text extraction from video
  using conditional random fields,'' in \emph{Document Analysis and Recognition
  (ICDAR), 2011 International Conference on}, Sept 2011, pp. 1029--1033.

\bibitem{Lu2010}
S.~Lu, B.~Su, and C.~L. Tan, ``Document image binarization using background
  estimation and stroke edges,'' \emph{International journal on document
  analysis and recognition}, pp. 1--12, 2010.

\bibitem{Sauvola1997}
J.~Sauvola, T.~Seppanen, S.~Haapakoski, and M.~Pietikainen, ``Adaptive document
  binarization,'' in \emph{Document Analysis and Recognition, 1997.,
  Proceedings of the Fourth International Conference on}, vol.~1.\hskip 1em
  plus 0.5em minus 0.4em\relax IEEE, 1997, pp. 147--152.

\bibitem{Sauvola2014}
G.~Lazzara and T.~G{\'e}raud, ``Efficient multiscale sauvola’s
  binarization,'' \emph{International Journal on Document Analysis and
  Recognition (IJDAR)}, vol.~17, no.~2, pp. 105--123, 2014.

\bibitem{Su2012}
B.~Su, S.~Lu, and C.~L. Tan, ``A learning framework for degraded document image
  binarization using markov random field,'' in \emph{Pattern Recognition
  (ICPR), 2012 21st International Conference on}.\hskip 1em plus 0.5em minus
  0.4em\relax IEEE, 2012, pp. 3200--3203.

\bibitem{EoE2012}
R.~F. Moghaddam, F.~F. Moghaddam, and M.~Cheriet, ``Unsupervised ensemble of
  experts (eoe) framework for automatic binarization of document images,'' in
  \emph{Document Analysis and Recognition (ICDAR), 2013 12th International
  Conference on}.\hskip 1em plus 0.5em minus 0.4em\relax IEEE, 2013, pp.
  703--707.

\bibitem{Phase2014}
H.~Ziaei~Nafchi, R.~Farrahi~Moghaddam, and M.~Cheriet, ``Phase-based
  binarization of ancient document images: Model and applications,'' 2014.

\bibitem{LTP2010}
X.~Tan and B.~Triggs, ``Enhanced local texture feature sets for face
  recognition under difficult lighting conditions,'' \emph{Image Processing,
  IEEE Transactions on}, vol.~19, no.~6, pp. 1635--1650, 2010.

\bibitem{ramirez2010transition}
M.~A. Ram{\'\i}rez-Orteg{\'o}n, E.~Tapia, L.~L. Ram{\'\i}rez-Ram{\'\i}rez,
  R.~Rojas, and E.~Cuevas, ``Transition pixel: A concept for binarization based
  on edge detection and gray-intensity histograms,'' \emph{Pattern
  Recognition}, vol.~43, no.~4, pp. 1233--1243, 2010.

\bibitem{sklearn}
F.~Pedregosa, G.~Varoquaux, A.~Gramfort, V.~Michel, B.~Thirion, O.~Grisel,
  M.~Blondel, P.~Prettenhofer, R.~Weiss, V.~Dubourg, J.~Vanderplas, A.~Passos,
  D.~Cournapeau, M.~Brucher, M.~Perrot, and E.~Duchesnay, ``Scikit-learn:
  Machine learning in {P}ython,'' \emph{Journal of Machine Learning Research},
  vol.~12, pp. 2825--2830, 2011.

\bibitem{geurts2006extremely}
P.~Geurts, D.~Ernst, and L.~Wehenkel, ``Extremely randomized trees,''
  \emph{Machine learning}, vol.~63, no.~1, pp. 3--42, 2006.

\bibitem{6482566}
T.~Lelore and F.~Bouchara, ``Fair: A fast algorithm for document image
  restoration,'' \emph{Pattern Analysis and Machine Intelligence, IEEE
  Transactions on}, vol.~35, no.~8, pp. 2039--2048, Aug 2013.

\bibitem{nafchi2013historical}
H.~Z. Nafchi, R.~F. Moghaddam, and M.~Cheriet, ``Historical document
  binarization based on phase information of images,'' in \emph{Computer
  Vision-ACCV 2012 Workshops}.\hskip 1em plus 0.5em minus 0.4em\relax Springer,
  2013, pp. 1--12.

\end{thebibliography}
\end{document}